\pdfoutput=1
\documentclass[pmlr]{jmlr}

 \usepackage{rotating}

\usepackage{booktabs}
\usepackage[load-configurations=version-1]{siunitx} 

\usepackage{pgf}
\DeclareUnicodeCharacter{2212}{-}
\usepackage[hang,flushmargin]{footmisc}

\makeatletter
\def\set@curr@file#1{\def\@curr@file{#1}} 
\makeatother

\theorembodyfont{\upshape}
\theoremheaderfont{\scshape}
\theorempostheader{:}
\theoremsep{\newline}

\jmlrvolume{149}
\jmlryear{2021}
\jmlrworkshop{Machine Learning for Healthcare}

\title[Deep Kernel Accelerated Failure Time Models]{Uncertainty-Aware Time-to-Event Prediction using Deep Kernel Accelerated Failure Time Models}

\author{\Name{Zhiliang Wu}
      \Email{zhiliang.wu@siemens.com}\\
      \addr Ludwig Maximilians University Munich\\
      Siemens AG, Technology, Munich
      \AND
      \Name{Yinchong Yang}
      \Email{yinchong.yang@siemens.com}\\
      \addr
      Siemens AG, Technology, Munich
      \AND
      \Name{Peter A. Fasching}
      \Email{peter.fasching@uk-erlangen.de}\\
      \addr Department of Gynecology and Obstetrics\\
      University Hospital Erlangen, Erlangen
      \AND
      \Name{Volker Tresp}
      \Email{volker.tresp@siemens.com}\\
      \addr Ludwig Maximilians University Munich\\
      Siemens AG, Technology, Munich
      }

\begin{document}

\maketitle

\begin{abstract}

Recurrent neural network based solutions are increasingly being used in the analysis of longitudinal Electronic Health Record data. However, most works focus on prediction accuracy and neglect prediction uncertainty. We propose Deep Kernel Accelerated Failure Time models for the time-to-event prediction task, enabling uncertainty-awareness of the prediction by a pipeline of a recurrent neural network and a sparse Gaussian Process. Furthermore, a deep metric learning based pre-training step is adapted to enhance the proposed model. Our model shows better point estimate performance than recurrent neural network based baselines in experiments on two real-world datasets. More importantly, the predictive variance from our model can be used to quantify the uncertainty estimates of the time-to-event prediction: Our model delivers better performance when it is more confident in its prediction. Compared to related methods, such as Monte Carlo Dropout, our model offers better uncertainty estimates by leveraging an analytical solution and is more computationally efficient.
\end{abstract}

\section{Introduction}
Since the introduction of the Electronic Health Record (EHR), an exploding amount of healthcare-related data has been collected in clinics. The physicians often become overwhelmed by data volume and data complexity and may turn to data-driven clinical decision support systems \citep{halpern2016electronic, tresp2016going, xiao2018opportunities}.
These solutions often provide decision support in two ways. A \emph{prescriptive} system generates action recommendations, such as medications and therapy plans, while a \emph{predictive} system provides physicians with a prediction of the outcome, given a decision. Such outcome could be, for instance, the adverse events related to a specific therapy or the progression-free-survival time after treatment. These predictions are often based on modeling the three-way interaction between the outcome, the patients' status, and clinical decisions recorded in historical data.
In this work, we address the prediction of treatment outcome and propose a new class of uncertainty-aware models that can communicate uncertainty to physicians. We argue that such uncertainty estimates add transparency and trustworthiness to the clinical decision support systems and encourage their application on even larger scales.

Due to the high-dimensional, sparse, and sequential nature of EHR data, simpler, more transparent white-box models often fail to capture the complex interactions between the target variable and input features.
Meanwhile, recurrent neural network (RNN)-based solutions have proven capable of addressing the longitudinal aspect in EHR data \citep{esteban2016predicting, choi2017using, yang2017predictive, purushotham2018benchmarking, wu2020learning}. These models apply RNNs to aggregate historical observations to produce an individual and time-dependent representation of a patient. The last layer is then a linear map from patient representation to the target variable representing, e.g., the predicted outcome for the patient. The advantage of an RNN is twofold. First, it can handle records that vary in length from patient to patient, as in the analysis of texts with varying lengths \citep{mikolov2012statistical}. Second, the more advanced RNN variants such as long short-term memory (LSTM) \citep{hochreiter1997long} and gated recurrent unit (GRU) \citep{chung2014empirical} are flexible in memorizing both long-term and short-term input features.
Despite the state-of-the-art predictive performance of the neural network (NN)-based methods, these methods fail to provide reasonable uncertainty estimates of the predictions \citep{nguyen2015deep}. It is easier to address this issue in classification tasks, as the predicted probability can be interpreted to reflect uncertainty, which leads to various calibration approaches, like temperature scaling \citep{guo2017calibration}. However, for regression tasks---like the time-to-event prediction task in our case---one has to provide a predictive distribution to address the uncertainty estimates, which is rarely considered in vanilla NN-based solutions.
In healthcare applications, we would argue that uncertainty-awareness of the model is as important as point estimate performance, since the uncertainty estimates would assist the physicians in better interpreting the results from a black-box model \citep{begoli2019need}. If the model provides a high uncertainty estimate for its prediction, the physicians would be more careful about it and take a closer look at that case.

It is, however, not a trivial task to augment NNs with reliable uncertainty estimates. Currently, popular choices to do so include MC Dropout methods \citep{gal2016dropout}, Bayesian neural networks \citep{bishop2006pattern} and deep ensembles \citep{lakshminarayanan2017simple}. Many variants of them involve repeated sampling procedures either during training or during inference, which could become computationally expensive for large NNs. In this work, we investigate the possibility of quantifying the predictive uncertainty by applying Gaussian Processes (GPs). As a popular class of machine learning methods, GP produces a predictive distribution instead of a single point estimate for each test sample. For small datasets,  GPs have proven to be flexible and data-efficient. However, GPs' inclusion of a large training dataset inevitably introduces large storage and computational complexity \citep{gpml}. With the recent advance of sparse GP techniques, the computational complexity has been largely reduced \citep{liu2020gaussian}. Motivated by the desirable predictive distribution of GPs and the progress of sparse GPs, we propose in this work a novel approach by integrating RNNs as feature extractors into GP-based predictive models. Furthermore, we propose a deep metric learning (DML)-based pre-training method to further improve the performance of the model.

In the context of time-to-event prediction, our proposed model is closely related to the Accelerated Failure Time (AFT) models \citep{prentice1978linear, kalbfleisch2002statistical}. As one of the well-known models in survival analysis, the AFT models have shown promising results, especially when, in an application, the direct prediction of survival time is more important than hazard estimation. With our proposed method, we have augmented the uncertainty estimates in the AFT models.

\subsection*{Generalizable Insights about Machine Learning in the Context of Healthcare}

Neural networks offer powerful modeling ability to learn from EHR data, but often neglect the uncertainty estimates in the predictions. Leveraging state-of-the-art sparse GPs, we propose a method to integrate the uncertainty into the NN-based solutions for time-to-event prediction tasks. Experiments on two real-world datasets show that the resulting model can 1) scale very well to large datasets; 2) deliver improved performance regarding point estimates; 3) offer reasonable predictive variances, which reflect the confidence of the predictions and enhance the calibration of the model. The uncertainty estimates in our model can help establish a trustworthy relationship with physicians since it expresses higher confidence in more accurate predictions and vice versa.

\section{Related Work}\label{sec: related}

\paragraph{Predictive Modeling with EHRs}
To better capture the time-dependent information of patients, many works have been proposed to model longitudinal EHR data, ranging from earlier statistical methods like \emph{landmarking} \citep{van2007dynamic}, \emph{Joint Models} \citep{rizopoulos2011dynamic, hickey2016joint}, to more recent methods like \emph{Bayesian Nonparametric Dynamic Survival} \citep{bellot2020flexible}.
Meanwhile, RNN-based approaches have proved to be very successful both for discrete medical events and continuous time-series data. \citet{esteban2016predicting} applied sequence-to-sequence RNN models to predict discrete medical events of patients suffering from kidney failure. \citet{yang2017predictive} proposed many-to-one RNN models to deal with discrete medical events and predict the therapy decision for breast cancer. At the same time, \citet{choi2017using} applied models of similar structure for the early detection of heart failure onset. For continuous time-series data, multiple readings of individual signals are usually aggregated to reduce the high-resolution patient data so that they can be better consumed by the neural networks, e.g., heart rate and blood pressure in ICU time-series data \citep{johnson2016mimic}. With the aggregation method on the ICU time series data, \citet{purushotham2018benchmarking} provided RNN-based benchmarks for the mortality prediction, length-of-stay prediction, and ICD-9 code group prediction. Following similar data pre-processing steps, \citet{wu2020learning} presented RNN-based models to learn the optimal treatment strategies for administering intravenous fluids and vasopressors. In this work, we include EHR data with discrete medical events as well as the dataset with continuous time-series measurements for the time-to-event prediction tasks to validate our proposed model.

\paragraph{Survival Analysis with Neural Networks and Gaussian Processes}
As most popular approaches in survival analysis are based on generalized linear models,
they have been extended to nonlinear models, including NNs and GPs. \citet{saul2016gaussian} proposed chained GPs to model multiple parameters of the log-logistic likelihood in AFT models through the latent functions of GPs. In addition, many GP-based methods are proposed to enhance the Cox proportional hazards (CPH) model, including the Bayesian semi-parametric model \citep{fernandez2016gaussian} and deep
multi-task Gaussian process DMGP \citep{alaa2017deep}.
For NN-based approaches, \citet{yang2017modeling} combines tensorized RNN model with the AFT model to predict progression-free survival (PFS) time. \citet{kvamme2019time} proposes an extension of CPH models, \emph{CoxTime}, by parametrizing the relative risk function with NNs as well as modeling interactions between covariates and time.
However, most of these works focus on capturing the non-linearity between covariates to improve the performance of point estimates. The uncertainty perspective of the prediction is rarely addressed. More recently, \citet{chen2020deep} proposed Deep Kernel Survival Analysis to learn kernel functions for the conditional Kaplan-Meier estimator and enables subject-specific survival time prediction intervals. The uncertainty is quantified by the prediction intervals. With an emphasis on uncertainty-awareness, we explore in this work the time-to-event prediction tasks with the AFT models using a combination of RNNs and GPs.

\paragraph{Exact Gaussian Process and Scalable Variational Gaussian Process with Neural Networks}
Since the capacity of GPs grows with available training data, many works have proposed possible solutions for both exact GP and sparse GP. Based on the efficient GP inference using Blackbox Matrix-Matrix multiplication from \citet{gardner2018gpytorch}, \citet{wang2019exact} realized exact GP training on over a million training samples by taking advantage of multi-GPU parallelization. Meanwhile, various works have been proposed to approximate the original GPs to save computational cost, including some early efforts, such as the Bayesian committee machine (BCM) \citep{tresp2000bayesian}, the Nyström methods \citep{williams2001using}, the Fully Independent Training Conditional (FITC) Approximation \citep{snelson2006sparse}, Variational Free Energy (VFE) \citep{titsias2009variational}, the more recent Scalable Variational Gaussian Process (SVGP) \citep{hensman2013gaussian} and Parametric Predictive Gaussian Process (PPGP) Regressor \citep{jankowiak2020parametric}
 (more details see Sec. \ref{sec: svgp_tp}). In addition, the idea of combining sparse GPs with neural networks also received much attention, where the Deep Kernel Learning (DKL) \citep{wilson2016deep} and GP hybrid deep networks (GPDNN) \citep{bradshaw2017adversarial} are the ones most related to our work.

\section{Methods}\label{sec: methods}
This section first discusses the RNN-based feature extractors to learn representations from the patients' static and sequential information. The resulting latent representations are used as inputs for the time-to-event prediction task. We will elaborate on our proposed model, which combines the GP-based models with AFT models. Afterward, a DML-based supervised pre-training method is presented to enhance the performance of the proposed model.

\subsection{Recurrent Neural Networks as Feature Extractors}\label{sec: rnnfe}

EHR data typically consist of \emph{static features} and \emph{sequential features}, both of which are important for the time-to-event prediction tasks. We regard the background information of each patient as static features $\boldsymbol{x}_i^{\text{sta}}\in \mathbb{R}^{\text{n}_{\text{sta}}}$. We use $i, \text{n}_{\text{sta}}$ to denote the patient sample index and the number of static features, respectively. In addition, features observed at all time-steps constitute the sequential feature matrix $\boldsymbol{X}_i^{\text{seq}}=[\boldsymbol{x}_i^0, \boldsymbol{x}_i^1, \dots, \boldsymbol{x}_i^{t_i}]^{\top} \in \mathbb{R}^{t_i \times \text{n}_{\text{seq}}}$, where $t_i$ is the number of observed time-steps for the $i$-th patient sample and $\text{n}_{\text{seq}}$ denotes the number of sequential features. Due to the high sparsity or redundancy in raw features spaces, it has been shown to be beneficial to first apply a (non-linear) embedding layer on the raw features to learn the static hidden representation $\boldsymbol{h}_i^{\text{sta}}$ and sequential feature embeddings $\boldsymbol{X}_i^{\text{seq\_emb}}$ \citep{esteban2016predicting}. Formally, we have
\begin{align*}
    \boldsymbol{h}_i^{\text{sta}} & = g_1(\boldsymbol{A} \boldsymbol{x}_i^{\text{sta}}) \in \mathbb{R}^{\text{n}_{\text{sta\_repr}}} \\
    \boldsymbol{X}_i^{\text{seq\_emb}} & = g_2(\boldsymbol{X}_i^{\text{seq}}\boldsymbol{B}) \in \mathbb{R}^{t_i \times \text{n}_{\text{seq\_emb}}}
\end{align*}
where $\boldsymbol{A}\in\mathbb{R}^{\text{n}_{\text{sta\_repr}}\times \text{n}_{\text{sta}}}, \boldsymbol{B}\in\mathbb{R}^{\text{n}_{\text{seq}}\times \text{n}_{\text{seq\_emb}}}$ are embedding matrices,  $g_1(\cdot), g_2(\cdot)$ are activation functions like $\tanh(\cdot)$, $\text{n}_{\text{sta\_repr}}, \text{n}_{\text{seq\_emb}}$ denote the dimension of the static hidden representations and the sequential feature embeddings, respectively. Afterward, more advanced variants of RNNs, LSTM or GRU, are used to encode the sequential feature embeddings $\boldsymbol{X}_i^{\text{seq\_emb}}$ into sequential latent representations $\boldsymbol{h}^{\text{seq}}_i$. Since we are mainly interested in modeling the time-to-event, only the last hidden states from LSTM/GRU are involved as inputs in the down-streaming tasks. Formally, we have
\begin{equation*}
    \boldsymbol{h}^{\text{seq}}_i = \text{RNN}(\boldsymbol{X}_i^{\text{seq\_emb}}) \in \mathbb{R}^{\text{n}_\text{seq\_repr}},
\end{equation*}
where $\text{n}_\text{seq\_repr}$ is the dimension of sequential hidden representations and $\text{RNN}(\cdot)$ could be an LSTM or GRU.
\begin{figure}[!b]
    \centering
    \includegraphics{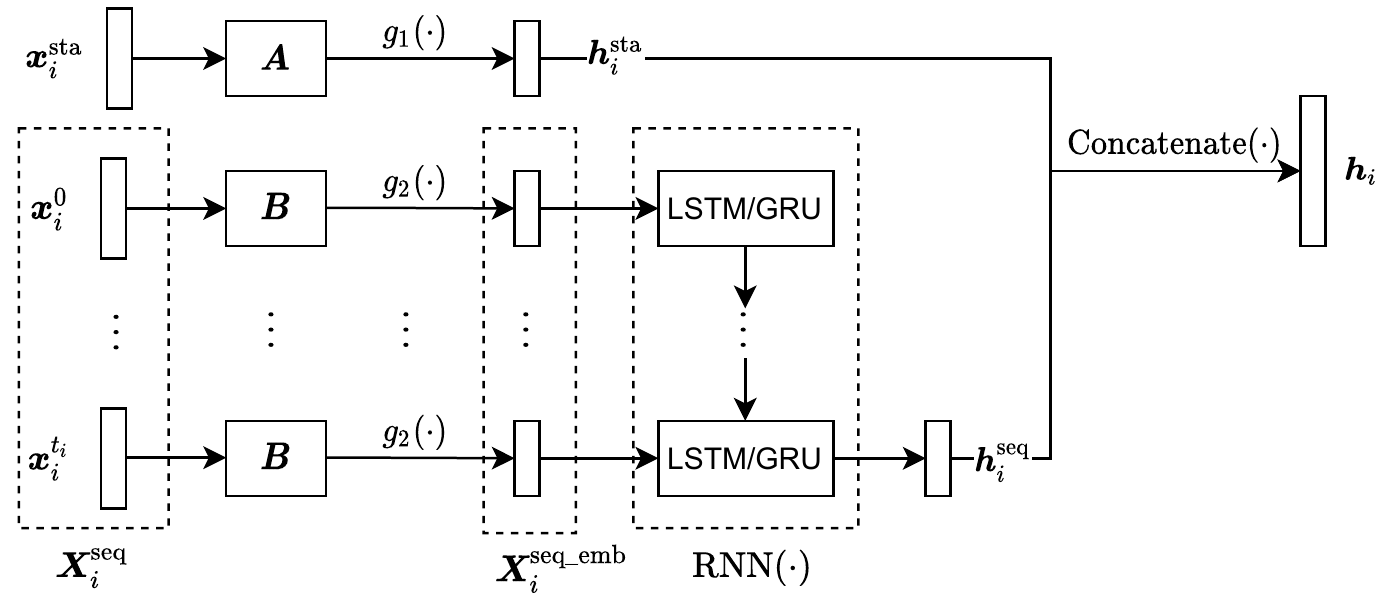}
    \caption{Illustration of the RNN-based feature extractor: An (non-linear) embedding layer is first involved in learning the sequential feature embeddings, which are then fed into RNN-based models to encode the sequential hidden representations. These are then concatenated with the static hidden representations. The complete hidden representation is expected to encode all relevant patient information and serves as abstract covariates for the down-streaming time-to-event prediction task. }
    \label{fig:rnnfe}
\end{figure}

 The complete structure of the feature extractor is shown in Fig.~\ref{fig:rnnfe}. Similar to the encoder part in \citet{cho2014learning}, this procedure is especially appealing for the patients with different observed time-steps, as it is theoretically capable of storing all relevant information in the medical events with variable lengths but remains a consistent form of representations.

\subsection{Scalable Variational Gaussian Processes for Time-to-Event Prediction}\label{sec: svgp_tp}
From the last section, we have learned the static hidden representation $\boldsymbol{h}_i^{\text{sta}}$ and the sequential hidden representation $\boldsymbol{h}_i^{\text{seq}}$ through RNN-based feature extractors. By concatenating them, we get the complete hidden representation as
$$
\boldsymbol{h}_i = [\boldsymbol{h}_i^{\text{sta}}; \boldsymbol{h}_i^{\text{seq}}]\in\mathbb{R}^{\text{n}_\text{sta\_repr}+\text{n}_\text{seq\_repr}},
$$
which can be viewed as abstract covariates of patients in a latent feature space.

The class of Accelerated Failure Time (AFT) models is a general class of models, where the covariates of the patients are assumed to act multiplicatively on the time-scale \citep{collett2015modelling}. Compared to the semi-parametric Cox proportional hazards (CPH) models, the AFT models take advantage of their parametric nature and include a wider range of survival time distributions.
Formally, with the time-to-event target variable $z_i$, the AFT models predict its logarithm as $\log z_i := y_i = \boldsymbol{\beta}^\top  \boldsymbol{h}_i + \epsilon_i ,
$ where $\boldsymbol{\beta}^\top \boldsymbol{h}_i$ is the linear predictor with the (trainable) parameter vector $\boldsymbol{\beta}$, $\epsilon_i \stackrel{i.i.d.}{\sim} \mathcal{D_\epsilon}$ denotes the error term which is specified by a particular probability distribution $\mathcal{D_\epsilon}$. Common choices for $\mathcal{D_\epsilon}$ include Normal, Weibull and Logistic distributions, which correspondingly specifies the target variable $z_i$ to be log-normal, log-weibull and log-logistic distributed, respectively. In this work, we assume our target variable of interest to follow a log-normal distribution. In other words,  we have correspondingly $\epsilon_i\stackrel{i.i.d.}{\sim} \mathcal{N}(0, \sigma^2_{\text{obs}})$. 
Furthermore, we propose to replace the linear predictor $\boldsymbol{\beta}^\top \boldsymbol{h}_i$ with Gaussian Process posterior prediction to enable uncertainty-aware predictions. In the following, we shall introduce this approach in detail.

\begin{figure}[!b]
    \centering
    \includegraphics{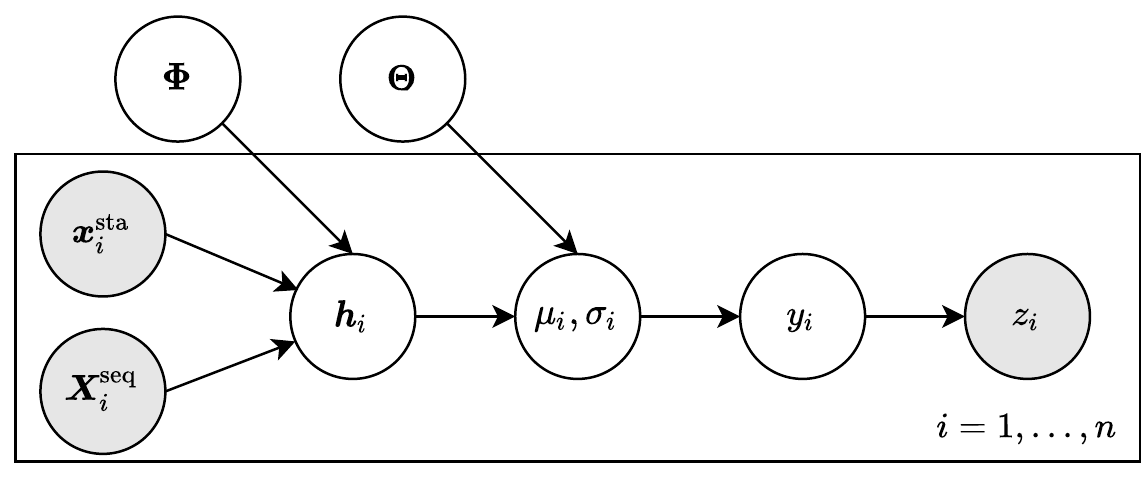}
    \caption{Graphical model of Deep Kernel Accelerated Failure Time models in plate notation. Nodes represent variables, where shaded ones are observable and non-shaded ones are latent. Plates indicate the repetition of the subgraph. }
    \label{fig: dkaft}
\end{figure}
As shown in Fig.~\ref{fig: dkaft}, after we obtain the hidden representations $\boldsymbol{h}_i$ from the feature extractor $f_{\boldsymbol{\Phi}}(\cdot)$ (details see Sec.~\ref{sec: rnnfe}), these are used as abstract patient covariates for the subsequent GP-based models $g_{\boldsymbol{\Theta}}(\cdot)$ to generate predictive distribution $y_i \sim \mathcal{N}(\mu_i, \sigma^2_i)$, the logarithm of the target variable. We denote $\boldsymbol{\Phi}$ and $\boldsymbol{\Theta}$ as the trainable parameters in the RNN-based feature extractor and the GP-based predictive model, respectively. Since we take advantage of neural networks with GP-based models as an advanced version of AFT models, we name it Deep Kernel Accelerated Failure Time (DKAFT) models. In the following, we will discuss different GP-based models in our proposed method.

In regression, $y_i$ is a noisy observation of the GP function value $f_i=f(\boldsymbol{h}_i)$, which is assumed to behave a priori according to
$$
p(\boldsymbol{f}|\boldsymbol{h}_1, \dots,\boldsymbol{h}_n)  = \mathcal{N}(\boldsymbol{0}, \boldsymbol{K}),
$$
where $\boldsymbol{f}=[f_1, \dots, f_n]^\top\in \mathbb{R}^n$ is a vector of GP function values and $\boldsymbol{K}\in\mathbb{R}^{n\times n}$ is a covariance matrix, whose entries are given by the covariance function $k_{ij}=k(\boldsymbol{h}_i, \boldsymbol{h}_j)$. The choice of the covariance function reflects the prior knowledge of the generative process of the model, where a Radial Basis Function (RBF) kernel is commonly used. There are some important hyper-parameters in the covariance function, e.g., the length-scale and signal variance in the RBF kernel, which can be learned through maximizing the log marginal likelihood defined as
\begin{equation}\label{loss: exactgp}
\mathcal{L}_{\text{ExactGP}}=-\frac{1}{2} \boldsymbol{y}^{\top}\left(\boldsymbol{K}+\sigma_{\text{obs}}^{2} I\right)^{-1} \boldsymbol{y}-\frac{1}{2} \log \left|\boldsymbol{K}+\sigma_{\text{obs}}^{2} I\right|-\frac{n}{2} \log 2 \pi.
\end{equation}
With the optimized parameters, the prediction of a test sample $f_*$ can be understood as computing the conditional probability of the test location given all values in the training dataset. Formally, the GP model outputs a predictive distribution as
\begin{equation}\label{inf: exactgp}
    f_*  \sim \mathcal{N}(\boldsymbol{k}^{\top}_{*}\left(\boldsymbol{K}+\sigma_{\text{obs}}^{2} \boldsymbol{I}\right)^{-1} \boldsymbol{y}, k_{**}-\boldsymbol{k}^{\top}_{*}\left(\boldsymbol{K}+\sigma_{\text{obs}}^{2} \boldsymbol{I}\right)^{-1} \boldsymbol{k}_{*}),
\end{equation}
where $\boldsymbol{k}_{*}=[k(\boldsymbol{h}_1, \boldsymbol{h}_*), \dots, k(\boldsymbol{h}_n, \boldsymbol{h}_*)]^\top \in \mathbb{R}^n$ denotes the covariance function values between the training inputs and the test input $\boldsymbol{h}_*$.
Please note that, in contrast to the original AFT formulation, the covariates $\boldsymbol{h}_i$ do not influence the logarithm of the target variable directly in GP. Instead, the accelerating effect is realized via the covariance function.

Equation~\ref{loss: exactgp} and Equation~\ref{inf: exactgp} reveal the training and inference step for our proposed DKAFT model with an \emph{Exact GP output layer}. However, the Exact GP cannot scale well to a large-scale dataset due to the $\mathcal{O}(n^3)$ computational complexity from the inverse operations of the large covariance matrix $\boldsymbol{K}$.

A tremendous amount of work has been proposed to address the scalability issue in the Exact GP, where the techniques of inducing points with variational inference have found most interest \citep{quinonero2005unifying}. In short, the inducing points constitute a ``summary" dataset, which is learned to generalize the original dataset to reduce the $\mathcal{O}(n^3)$ computational complexity. They consist of inducing inputs $\{\boldsymbol{u}_i\}_{i=1}^m=:\boldsymbol{U}$ (corresponding to $\{\boldsymbol{h}_i\}_{i=1}^n$) and inducing variables $\{v_i\}_{i=1}^m=:\boldsymbol{v}$ (corresponding to $\{ f_i\}_{i=1}^n$), where $m\ll n$. In the context of our DKAFT model, the inducing inputs refer to a summary of the abstract patient covariates in the latent space. The learning of the inducing points is facilitated by variational methods under different approximation assumptions, e.g., the prior approximation and posterior approximation \citep{liu2020gaussian}.

Among various GP approximations, the Scalable Variational Gaussian Process (SVGP) proposed by \citet{hensman2013gaussian} reduces the computational complexity to $\mathcal{O}(m^3)$ and makes the training amenable to stochastic gradient descent (SGD)-based methods. More concretely, the variational distribution $q(\boldsymbol{v})$ of the inducing variables is assumed to follow a multivariate Normal distribution $\mathcal{N}(\boldsymbol{m}, \boldsymbol{S})$ in SVGP. Following the notation in \citet{jankowiak2020parametric}, instead of the log marginal likelihood objective in an Exact GP, we optimize the parameters by maximizing the Evidence Lower BOund (ELBO)
\begin{equation}\label{loss: svgp}
\mathcal{L}_\text{SVGP}=
\sum_{i=1}^{n}\left\{\log \mathcal{N}\left(y_{i} \mid \mu_{\boldsymbol{f}}\left(\boldsymbol{h}_{i}\right), \sigma_{\text {obs }}^{2}\right)-\frac{\sigma^2_{\boldsymbol{f}}\left(\boldsymbol{h}_{i}\right)}{2 \sigma_{\text {obs }}^{2}}\right\} - \mathrm{KL}(q(\boldsymbol{v}) \| p(\boldsymbol{v} ))
\end{equation}
and the predictive distribution for each sample is
\begin{equation}\label{inf: svgp}
    f_i\sim  \mathcal{N}(\mu_{\boldsymbol{f}}(\boldsymbol{h}_i), \sigma^2_{\boldsymbol{f}}(\boldsymbol{h}_i)) = \mathcal{N}(\boldsymbol{k}_{i}^{\top}\boldsymbol{K}_{\boldsymbol{vv}}^{-1} \boldsymbol{m}, k_{ii}- \boldsymbol{k}_{i}^{\top} \boldsymbol{K}_{\boldsymbol{ vv}}^{-1} \boldsymbol{k}_{i} + \boldsymbol{k}_{i}^{\top} \boldsymbol{K}_{\boldsymbol{vv}}^{-1} \boldsymbol{S} \boldsymbol{K}_{\boldsymbol{vv}}^{-1} \boldsymbol{k}_{i}).
\end{equation}
$\boldsymbol{K}_{\boldsymbol{vv}}\in \mathbb{R}^{m\times m}$ is the covariance matrix of the inducing variables, whose entries are computed based on inducing inputs as $k^{\boldsymbol{vv}}_{ij}=k(\boldsymbol{u}_i, \boldsymbol{u}_j)$,  $\boldsymbol{k}_i=[k(\boldsymbol{u}_1, \boldsymbol{h}_i), \dots, k(\boldsymbol{u}_m, \boldsymbol{h}_i)]^\top \in \mathbb{R}^m$ is the covariance function values between all inducing inputs with the sample input $\boldsymbol{h}_i$, and $\mathrm{KL}(\cdot \| \cdot)$ denotes the Kullback–Leibler divergence between two distributions.

Both training and inference of SVGP in Equation~\ref{loss: svgp} and Equation~\ref{inf: svgp} are more computationally tractable, since they only involve the inducing points instead of the whole dataset as in Exact GP. We can therefore take advantage of this formulation for large-scale datasets. With $\mathcal{L}_\text{SVGP}$ as an objective, we have our DKAFT model with an \emph{SVGP output layer}.

More recently, \citet{jankowiak2020parametric} found that the predictive uncertainty from SVGP is dominated by the input-independent observational noise $\sigma_{\text{obs}}^2$, whereas it is indeed the input-dependent function variance $\sigma^2_{\boldsymbol{f}}(\boldsymbol{h}_i)$ that makes the GP posteriors attractive. Different from the SVGP objective in Equation~\ref{loss: svgp}, the Parametric Predictive Gaussian Process (PPGP) Regressor takes advantage of the predictive distribution in Equation~\ref{inf: svgp} and embeds it directly in the objective using Maximum Likelihood Estimation (MLE) methods. Formally, the objective in PPGP is defined as
\begin{equation}\label{loss: ppgp}
\mathcal{L}_\text{PPGP}
=\sum_{i=1}^{n} \log \mathcal{N}\left(y_{i} \mid \mu_{\boldsymbol{f}}\left(\boldsymbol{h}_{i}\right), \sigma_{\mathrm{obs}}^{2}+\sigma^2_{\boldsymbol{f}}\left(\boldsymbol{h}_{i}\right)\right) - \mathrm{KL}(q(\boldsymbol{v}) \| p(\boldsymbol{v})).
\end{equation}
With $\mathcal{L}_\text{PPGP}$ as a training objective, we have our DKAFT model with \emph{an PPGP output layer}.

The objectives introduced above are defined for samples with observed time-to-event. For right-censored cases, we can take advantage of the parametric predictive distribution in Equation~\ref{inf: svgp} to compute the survival function, whose logarithm contributes to the final objective together with the ELBO objective. Such an optimization objective is also used in AFT models, where the non-censored cases contribute to the objective through their respective probability distribution function and the censored ones through the survival function \citep[see][Equation 5.9]{collett2015modelling}. Formally, the survival function of the log-normal distribution in our DKAFT model is

$$
S(z |\boldsymbol{h_i}) = 1 - \Phi\bigg(\frac{\log{z}-\mu_{\boldsymbol{f}}(\boldsymbol{h}_i)}{\sigma_{\boldsymbol{f}}(\boldsymbol{h}_i)+\sigma_{\text{obs}}}\bigg),
$$

where $\Phi(\cdot)$ is the cumulative distribution function of a standard normal distribution.

\subsection{Deep Metric Learning as Supervised Pre-training}
The proposed architecture with RNNs as feature encoders and sparse GPs as predictive models is trainable in an end-to-end fashion with gradient descent. The free parameters include parameters in RNNs, the inducing points, and hyper-parameters in the covariance function. In our experiment, we realize that training such architecture from scratch could be challenging. Given an RNN and inducing points that are both randomly initialized, the covariance matrix in GP is also random. This often causes the length scale parameter in the RBF kernel to shrink to extremely small values, and the GP would then degrade to its prior, correspondingly.
To alleviate such problems, we find the initialization of the inducing points to be an important step for obtaining good models. More specifically, we find that the training always fails if we initialize the inducing inputs with random vectors. On the other hand, initializing the inducing points with latent representations from the RNN-based feature extractor always shows good performance, even though the parameters in the feature extractor are initialized randomly. Formally, we get the initial inducing inputs as
$$
\boldsymbol{u}_i^{\text{init}}=\boldsymbol{h}_i^{\text{init}} =f_{\boldsymbol{\Phi}_{\text{init}}}(\boldsymbol{x}_i^{\text{sta}}, \boldsymbol{X}_i^{\text{seq}}),
$$
where a random subset of the training inputs $\{\boldsymbol{x}_i^{\text{sta}}, \boldsymbol{X}_i^{\text{seq}}\}_{i=1}^m$ is involved.
To this end, we conjecture that a pre-training step on the feature extractor would boost the performance of our DKAFT model. Since many covariance functions, e.g., RBF kernels, take the distance between samples as input, it would be beneficial if the feature extractor generates abstract covariates in well-clustered latent spaces, where the samples with similar target variables are closer to each other. We propose that one could achieve such a beneficial configuration via Deep Metric Learning (DML), which is initially proposed for vision-related tasks like face verification \citep{schroff2015facenet} and person re-identification \citep{hermans2017defense}. What DML learns is to represent samples in a latent space that retains the similarity in the target variables.

In DML, pair loss or triplet loss provides the foundation for embedding samples using twin networks, which refers to the replications of the same feature extractor network. Various losses have been proposed to improve the embedding from different perspectives, including contrastive loss \citep{hadsell2006dimensionality}, triplet margin loss \citep{weinberger2006distance} or the more recent Signal-To-Noise Ratio loss \citep{yuan2019signal}.
More specifically, a triplet is defined with the class information to consist of an anchor, a positive, and a negative sample, $\{\boldsymbol{x}_i^{\text{sta}}, \boldsymbol{X}_i^{\text{seq}}\}^{\text{A/P/N}}$, where the anchor is of the same class as the positive and the negative is not. As there are no class labels in time-to-event prediction, we propose to categorize the target variables according to their binnings in the histogram to facilitate the triplet generation. In the context of our DKAFT model, we train the RNN-based feature extractor using, e.g., triplet margin loss (\citet{schroff2015facenet})
$$
    \mathcal{J}_\text{triplet}=\sum_{i=1}^n \big[\text{d}(\boldsymbol{h}^\text{A}_i, \boldsymbol{h}^\text{P}_i) -  \text{d}(\boldsymbol{h}^\text{A}_i, \boldsymbol{h}^\text{N}_i) + \alpha]_+,
$$
where $\text{d}(\cdot, \cdot)$ is a distance metric, like Euclidean distance, $\boldsymbol{h}^\text{A}_i, \boldsymbol{h}^\text{P}_i, \boldsymbol{h}^\text{N}_i$ are the abstract covariates of anchor, positive, and negative samples, $\alpha$ is a predefined margin value, and $[\cdot]_+$ takes the positive part of the variable.
From the GP perspective, it is the covariance function that defines the ``similarity" between samples, the choice of a specific loss in DML should therefore take it into consideration.

To find a suitable training epoch for the pre-training, we use an early stopping technique, which terminates the training automatically if the monitored metric does not improve over a given number of epochs. Mean Average Precision at R (MAP@R) proposed in \citet{musgrave2020metric} is used as the monitored metric on the validation set, which combines the metrics of Mean Average Precision and R-precision.

To conclude, we propose to apply an RNN-based feature extractor to learn fix-sized latent representations from patient trajectories of variable lengths. The feature extractor can be randomly initialized or pre-trained with our proposed DML-based approach. The DKAFT model is trained end-to-end against (sparse) GP objectives using SGD-based methods and produces predictive distributions.

\section{Cohort}\label{sec: cohort}

\subsection{Data Extraction}

We have included two datasets to validate the effectiveness of our proposed method. In both datasets, we treat observed time-to-event as our target variables.

The first dataset is provided by the PRAEGNANT study network \citep{fasching2015biomarker}, which focuses on patients suffering from metastatic and incurable breast cancer. Based on a patient's background information and medical history, we attempt to predict the Progression-Free Survival time (PFS-PRAEGNANT), i.e., the number of days till the next recorded progression. The raw data are hosted in a relational database system, secuTrial\textsuperscript{\tiny\textregistered}, and can be accessed under restrictions. After querying and preprocessing, we retrieved a dataset of 1336 patient cases.

The second dataset comes from the Medical Information Mart for Intensive Care database (MIMIC-III), a freely accessible database, which contains data including $53,423$ distinct Intensive Care Unit (ICU) admissions of adult patients between 2001 and 2012 \citep{johnson2016mimic}. In this work, we consider a cohort of patients from MIMIC-III v1.4, who are older than $15$ years at the time of ICU admission. Besides, only the first admission of these patients is included to prevent potential information leakage in the analysis. Based on the data collected during the first $48$ hours, we attempt to predict the length-of-stay (LoS-MIMIC) for each admission, i.e., the number of days between hospital admission and discharge from the hospital. More specifically, we followed the scripts\footnote{\url{https://github.com/USC-Melady/Benchmarking_DL_MIMICIII}} provided by \citet{purushotham2018benchmarking} and extracted a dataset with $31,986$ patient admissions.

In both extracted datasets, all cases are with observed time-to-event. In case of the MIMIC dataset, the patients were always supposed to leave the ICU and the PRAEGNANT patients all have metastasis and were expecting multiple progressions.

\subsection{Feature Processing}
In the PRAEGNANT dataset, the static information includes 1) basic patient information, e.g., age and height 2) information on the primary tumor, and 3) history of metastasis before entering the study. In total, there are 26 features of binary, categorical, or numerical type. After performing one-hot encoding on the binary and categorical features, we obtain a feature vector $\boldsymbol{x}_i^{\text{sta}}\in \mathbb{R}^{114}$ for the static information with an average sparsity of $0.871$. The sequential information includes 4) local recurrence 5) metastasis 6) clinical visits 7) radio-therapies 8) systemic therapies, and 9) surgeries. These events are observed with a timestamp, and multiple events could happen at the same timestamp. After performing binary-encoding on 26 sequential features, we extract a feature matrix $\boldsymbol{X}^\text{seq}_i\in \mathbb{R}^{t_i \times 188}$ for the sequential information of each patient case, where $t_i$ denotes the length of the sequence before the progression. The length of the sequences $t_i$ varies from $1$ to $22$ and is on average $6.42$. The average sparsity of the sequential feature matrix is $0.973$.

In the MIMIC-III dataset, the static information refers to the basic information during the admission, e.g., age and admission type. After performing binary encoding on five static features, we obtain a feature vector for each admission $\boldsymbol{x}^\text{sta}_i\in \mathbb{R}^{10}$. Moreover, the sequential information refers to the continuously monitored measurements or prescriptions in the ICU environment. They are sampled or aggregated every one hour to represent the patient status at different time steps. 136 sequential features have been selected. Those features are available for most patients. As a result, we extract a feature matrix $\boldsymbol{X}^\text{seq}_i\in \mathbb{R}^{t_i \times 136}$ for the sequential information, where $t_i$ is 48 for all patient admissions. A complete list of the chosen features can be found in Appendix~\ref{app: features}.

\section{Experiments}\label{sec: exp}

\subsection{Experimental Details and Evaluation Approaches}
We conducted cross-validations (CV) for the PFS-PRAEGNANT and LoS-MIMIC prediction tasks with $90\%$ samples in the dataset. The validation set is used for tuning hyper-parameters like, e.g., the dimension of the latent representations, weight decay, and training epochs. As a result, we have $4, 32, 128$ and $4, 64, 64$ for the size of the static latent representations $\text{n}_{\text{sta\_repr}}$, sequential feature embeddings $\text{n}_{\text{seq\_emb}}$, and sequential latent representations $\text{n}_{\text{seq\_repr}}$ in the PFS-PRAEGNANT and LoS-MIMIC prediction tasks, respectively. The evaluation metrics reported in the following are all computed based on the remaining unseen $10\%$ samples of the dataset.

All our NN-based models are built with the PyTorch package \citep{NEURIPS2019_9015}. The GP-related methods are implemented with the help of the GPyTorch package \citep{gardner2018gpytorch}.
Related scripts\footnote{\url{https://github.com/ZhiliangWu/DKAFT}} are published to ensure the reproducibility of the work.

Since the target variable $z_i$ is assumed to follow a log-normal distribution, it is not appropriate to measure the results using Root Mean Square Error (RMSE) in its original scale. Therefore, we report the more robust metric of Median Absolute Deviation (MAD) defined as $\text{MAD}=\text{median}_i(|z_i - \hat{z}_i|)$. In addition, we report the RMSE in the logarithmic scale of the target variable, which is defined as $\text{RMSE} = \sqrt{\frac{1}{n}\sum_i^n (y_i - \hat{y}_i)^2}$.

In addition to the point estimate performance, we also evaluate the meaningfulness of the predictive variance $\sigma_i$ of our proposed model as well as other uncertainty-aware baselines using a \emph{Quantile Performance (QP) plot} \citep{wu2021quantifying, yang2021multi}.
Intuitively, the predictive confidence generated by a model is only systematically meaningful if the model assigns higher confidence to the more accurate predictions and lower confidence to the less confident ones.
In the scope of our work, we interpret the predictive variance as a form of confidence estimation, where smaller values correspond to higher confidence. Therefore, the predictive variance from our model enables a formal evaluation of such expected behavior.
Concretely speaking, we extract the evaluation pairs $\{y_i, \hat{y}_i\}_{i=1}^{n}$ (or $\{z_i, \hat{z}_i\}_{i=1}^{n}$) for each quantile of the predictive variance of the model $q\in \{ \frac{1}{Q}, \frac{2}{Q}, \dots, 1\}$, where the corresponding predictive variance $\sigma^2_i$ is smaller than or equal to the $q$-th quantile.
Formally, we have the performance in each quantile as
$$
\text{Performance}_\text{q} := \text{Metric}(\{(y_i, \hat{y}_i) \text{ or } (z_i, \hat{z}_i)~|~\forall \sigma_i^2 \leq q\text{-th quantile} \}),
$$
where $\text{Metric}(\cdot)$ could be MAD for $z_i$ or RMSE for $y_i$. We plot the performance of each quantile on the $y$-axis against the corresponding $q$ values on the $x$-axis. For a model with meaningful uncertain-awareness, a monotonically increasing line is expected in the QP plot. Furthermore, a stronger correlation between the metric and confidence across the quantiles suggest a better quantification of the predictive uncertainty.

Finally, we plot the (empirical) Cumulative Distribution Function (CDF) of the normalized residuals to show how well our model is calibrated as a further evaluation metrics. According to our normal distribution assumption on the logarithmic scale of the target variable, it is expected to be close to the CDF of a standard normal distribution $\mathcal{N}(0, 1)$. In addition, the Continuous Ranking Probability Score (CRPS), a popular calibration metric for regression \citep{gneiting2007strictly, jankowiak2020parametric}, is reported to have a quantitative comparison.

\subsection{Evaluation of the PFS-PRAEGNANT and LoS-MIMIC prediction}
As weak baselines, we report the performance of standard Cox and AFT regression using the R package \textit{survival} \citep{survival-book, survival-package} with raw features aggregated w.r.t. the time axis.
Such aggregation has been used in \citet{esteban2015predicting} and \citet{yang2017modeling}, which turns out to be a reasonable solution to deal with features with time-stamps by ignoring the order of the events.

To investigate the performance of different output layers as we discussed in Sec.~\ref{sec: methods}, we train all models with the same RNN-based feature extractor. Using a linear output layer as our strong baseline (\textit{RNN+AFT}) \citep{yang2017modeling}, we include the SVGP and PPGP output layers for both prediction tasks, which are denoted as \textit{DKAFT (SVGP)} and \textit{DKAFT (PPGP)}, respectively. Thanks to the moderate size of the PRAEGNANT dataset,  we also have an ExactGP output layer for the PFS-PRAEGNANT prediction task, which we denote as \textit{DKAFT (ExactGP)}.
To validate our proposed initialization method, we pre-trained the feature extractor using DML and then fine-tune the proposed model. In such a case, the performance of models without pre-training naturally serves as the baselines.
Note that, for our DKAFT models, only the mean predictions are involved in the evaluation of point estimates.
Results are summarized in Tab.~\ref{tab: exp_res}. For NN-based CPH baselines \citep{kvamme2019time}, we report the results in Tab.~\ref{tab: exp_all} in Appendix~\ref{app: baselines}.

\begin{table}[t]
\small
\hspace{-0.75cm}
\begin{minipage}{\textwidth+1.5cm}
\renewcommand\footnoterule{}
\centering
\caption{Experimental results: MAD in the original scale (days) and RMSE in the logarithmic scale for PFS-PRAEGNANT and LoS-MIMIC prediction tasks. Our DKAFT models outperform the baselines, including Cox Regression, AFT regression, and RNN+AFT models. More baselines in Appendix.~\ref{app: baselines}. }\label{tab: exp_res}
\begin{tabular}{@{}cccccc@{}}
\toprule
             &             & \multicolumn{2}{c}{Progression-Free Survival}             & \multicolumn{2}{c}{Length-of-Stay}                      \\
Method & Pre-training & MAD                            & RMSE                        & MAD                        & RMSE                      \\ \midrule
Cox Regression         & $-$\footnote{Not applicable to the method.}      & $200.800 \pm 16.984$         & $1.609 \pm 0.054$ & $2.727 \pm 0.007$ & $0.638 \pm 0.0002$ \\
AFT Regression        & $-^*$      & $206.065 \pm 8.988$    & $1.685 \pm 0.080$ & $2.742 \pm 0.014$ & $0.630 \pm 0.0001$ \\
RNN+AFT       & None        & $150.918 \pm 3.009$          & $1.273 \pm 0.019$          & $2.476 \pm 0.040$          & $0.575 \pm 0.003$          \\
DKAFT (ExactGP)      & None        & $144.622 \pm 8.689$          & $1.225 \pm 0.022$          & $-$\footnote{Not possible due to the $\mathcal{O}(n^3)$ computational complexity.}                          & $-^\dagger$                          \\
DKAFT (SVGP)         & None        & $154.237 \pm 13.490$         & $1.211 \pm 0.020$          & $2.428 \pm 0.056$          & $0.572 \pm 0.003$          \\
DKAFT (PPGP)         & None        & $147.108 \pm 6.284$          & $1.220 \pm 0.019$          & $2.351 \pm 0.021$          & $0.563 \pm 0.001$          \\
RNN+AFT       & DML      & $138.155 \pm 7.496$          & $1.267\pm 0.007$           & $2.452 \pm 0.057$          & $0.568 \pm 0.001$          \\
DKAFT (ExactGP)      & DML      & $\mathbf{134.422 \pm 7.255}$ & $1.202 \pm 0.012$          & $-^\dagger$                          & $-^\dagger$                          \\
DKAFT (SVGP)         & DML      & $151.852 \pm 11.305$         & $1.221 \pm 0.007$          & $2.438 \pm 0.079$          & $0.567 \pm 0.005$          \\
DKAFT (PPGP)          & DML      & $146.616 \pm 17.109$         & $\mathbf{1.195 \pm 0.008}$ & $\mathbf{2.346 \pm 0.042}$ & $\mathbf{0.557 \pm 0.002}$ \\
\bottomrule
\end{tabular}
\end{minipage}
\end{table}

\begin{table}[b]
\begin{minipage}{\textwidth}
\renewcommand\footnoterule{}
\centering
\caption{$p$-values from paired t-tests to assess the significance of improvement achieved by our DKAFT models, based on the RMSEs collected from multiple cross validations. A two factor ANOVA on the effect of different model setups can be found in the Appendix.\ref{app: anova} }\label{tab:t-test}
\begin{tabular}{@{}ccccc@{}}
\toprule
                   & \multicolumn{2}{c}{No pre-training} & \multicolumn{2}{c}{DML} \\
Comparison candidates            & PFS            & LoS               & PFS        & LoS        \\ \midrule
DKAFT (ExactGP) vs. RNN+AFT & $0.021$        & $-$\footnote{Not possible due to the $\mathcal{O}(n^3)$ computational complexity.}                 & $2\text{e-}4$ & $-^*$           \\
DKAFT (SVGP) vs. RNN+AFT    & $0.010$        & $0.282$           & $0.001$ & $0.682$    \\
DKAFT (PPGP) vs. RNN+AFT    & $0.002$        & $0.001$        & $3\text{e-}4$  & $3\text{e-}4$ \\ \bottomrule
\end{tabular}
\end{minipage}
\end{table}

From Tab.~\ref{tab: exp_res} we can see that our DKAFT models demonstrate much stronger performance compared to the Cox Regression and AFT Regression with aggregated features. For the evaluation w.r.t. RMSE, our DKAFT models all outperform the corresponding strong baselines, the RNN+AFT models. Tab.~\ref{tab:t-test} shows the $p$-values of paired t-tests quantifying the performance improvement of our DKAFT models. In the task of PFS-PRAEGNANT prediction, we can see our DKAFT models outperform RNN+AFT models significantly (with a significant level $\alpha=0.05$). In contrast, only DKAFT (PPGP) models show significantly better RMSE performance in the LoS-MIMIC prediction task. For the evaluation regarding MAD, our DKAFT (ExactGP) model performs best in the PFS-PRAEGNANT prediction task, while it is DKAFT (PPGP) for the LoS-MIMIC prediction task. Meanwhile, we can observe a moderate improvement for most models pre-trained with DML compared to those trained from scratch.

As discussed in Sec.~\ref{sec: methods}, DKAFT models are trained by either optimizing the log marginal likelihood objective or the variational approximation of it, which is essentially a generalization to mean square error in linear regression. Therefore, improved performance is expected as the GP-based output layers include inducing points during the training and inference time, which implicitly facilitates integrating all possible linear models. Meanwhile, since MAD is a more robust metric than RMSE regarding outliers, there are some performance differences of the same model between these two metrics.

The superior performance of the DKAFT (SVGP) and DKAFT (PPGP) compared to DKAFT (ExactGP) w.r.t. RMSE is a bit surprising since these models are essentially approximations of the latter. We speculate that such phenomenon comes from the sparse and high-dimensional features in the PRAEGNANT dataset, which results in redundant or even repetitive patient covariates from the feature extractor. This makes the GP model struggle to capture the correlation correctly. On the contrary, the inducing points in SVGP or PPGP are not constrained by the raw input features and could avoid this coupling during the optimization.

For the initialization with pre-training methods, DML helps attain a feature extractor, which clusters samples with similar targets in nearby regions. Models with all output layers, including linear layers, achieve moderate improvement. This can be attributed to 1) the non-convexity nature of the log marginal likelihood objective or the ELBO objective with deep neural networks, where a good starting point could offer advantages for the following optimization procedure; 2) the good initialization of the inducing inputs generated by the pre-trained feature extractor.

\subsection{Evaluation of the predictive variances}
\begin{figure}[t]
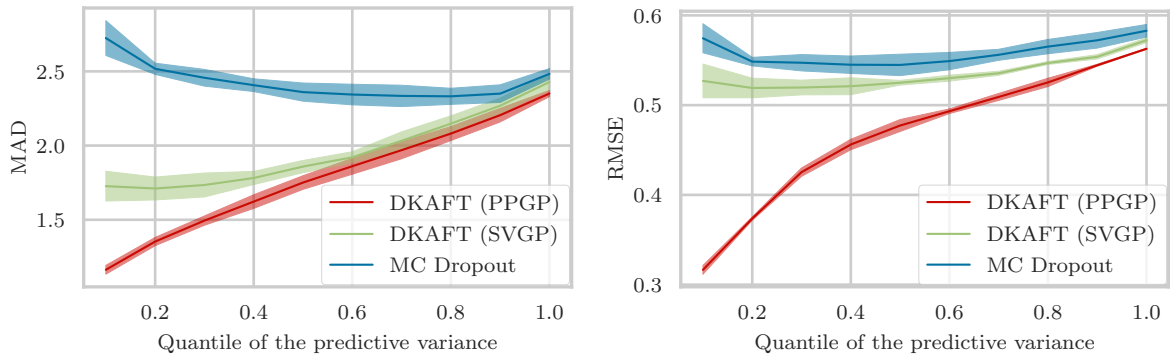

  \begin{minipage}{0.49\textwidth}
    \centering
    \input{qp_mad_50.pgf}
  \end{minipage}
  \quad
  \begin{minipage}{0.49\textwidth}
    \centering
    \input{qp_rmse_50.pgf}
  \end{minipage}
    \caption{Quantile MAD (left) and Quantile RMSE (right) in the y-axis against quantile predictive uncertainty in the x-axis for the LoS-MIMIC prediction task. Our DKAFT model with a PPGP output layer (red) shows the strongest increasing trend in both MAD and RMSE. This indicates that the model is monotonically more confident in predictions that are indeed closer to the ground-truths. We emphasize that this is a desirable feature to expect from an uncertainty-aware prediction model.}
  \label{fig: qp_plot}
\end{figure}

Apart from improved performance for point estimates, the main advantage of the proposed method lies in the uncertainty-aware nature of the model. As a baseline, we include MC Dropout \citep{gal2016dropout}, a well-known method for enabling uncertainty estimates in neural networks. By adding a dropout layer \citep{srivastava2014dropout} before each weight layer, the resulting model is proved to be mathematically equivalent to a probabilistic deep Gaussian Process. In our experiments, we followed the proposed method with the suggested dropout rate of $0.2$. The mean prediction and function variance are computed from performing $50$ stochastic forward passes through the network. Since QP plots demonstrate unstable performance if the number of evaluation pairs in each quantile is too small (like the test set for the PFS-PRAEGNANT prediction task with only 134 samples in total), we only report the evaluation on the LoS-MIMIC prediction task in Fig.~\ref{fig: qp_plot}.

In Fig.~\ref{fig: qp_plot} we visualize the quantile performance for MAD and RMSE, where the solid line represents an average performance and the error bar for the standard deviation across CV splits. We observe a strong monotonically increasing line in both metrics from our DKAFT (PPGP) model. For the evaluation pairs that the model is more confident with, e.g., ones corresponding to the predictive variances at quantile $10\%$, the MAD and RMSE are $1.163 \pm 0.030$ and $0.316 \pm 0.005$, respectively. Such results correspond to only half of the values reported in Tab.~\ref{tab: exp_res}, indicating a significant improvement. Meanwhile, the DKAFT (SVGP) model shows an increasing dependency only in MAD. For the models with MC Dropout, there seems to be no performance difference between quantiles.

Compared with MC Dropout, our DKAFT models deliver more meaningful uncertainty estimates since the inducing point technique realizes explicit modeling of the predictive variances.
As expected, the most meaningful uncertainty estimates are visible from the models with a PPGP output layer, since the function variance is restored explicitly in the training objective compared to those with an SVGP output layer. These observations indeed motivated the application of our DKAFT (PPGP) model when meaningful uncertainty estimates become a higher priority.

\subsection{Calibration of the Model}
\begin{figure}[t]
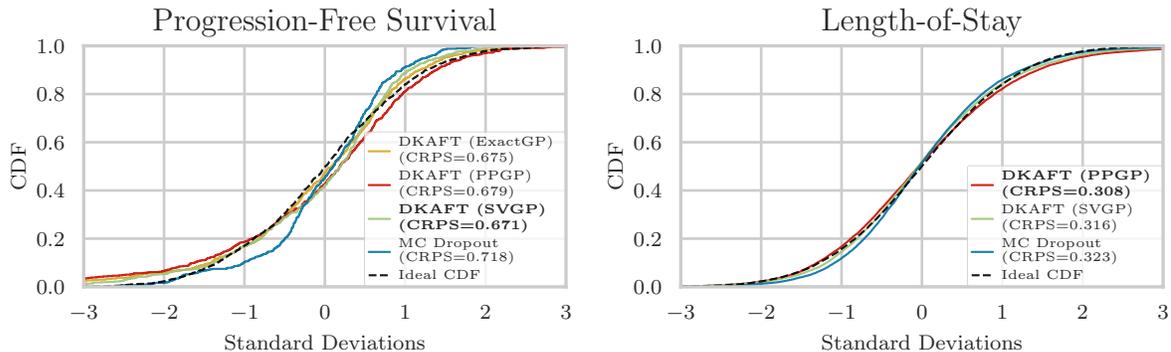

  \begin{minipage}{0.49\textwidth}
    \centering
    \input{ecdf_fasching_.pgf}
  \end{minipage}
  \quad
  \begin{minipage}{0.49\textwidth}
    \centering
    \input{ecdf_setc_.pgf}
  \end{minipage}
    \caption{Empirical CDF of the normalized residual, $(y_i - \mu_i )/ \sigma_i $, from different models against an "ideal CDF" from a standard normal distribution. Continuous Ranking Probability Score is reported to provide a quantitative comparison. All models show well-calibrated behavior, where our DKAFT models are better calibrated than the baseline method, MC Dropout. }
  \label{fig: ecdf_plot}
\end{figure}

Calibration of a predictive model refers to the statistical consistency between the predictive distribution from the model and the observations \citep{gneiting2007probabilistic}, which is arguably also an important aspect for healthcare applications.
In GPs, the predictive variance incorporates both the modeling uncertainty (function variance) and data uncertainty (observational noise). Even for data-points lying far from the training data, the resulting predictive distribution tends to be well-calibrated \citep{gpml}. In this section, we demonstrate that our DKAFT models inherit this nice property from GPs.
Meanwhile, as a popular method for calibrating neural networks, MC Dropout is included in our experiment as a baseline.
We visualize the empirical CDF of the normalized residuals with the predictive variances in Fig.~\ref{fig: ecdf_plot}. Besides, the CRPS score is computed to have a quantitative comparison. The CRPS score generalizes the Mean Absolute Error (MAE) to probabilistic predictions.

In Fig.~\ref{fig: ecdf_plot}, we visualize the CDF plots for both time-to-event prediction tasks. Graphically speaking, all methods demonstrate well-calibrated behavior based on their closeness to the best possible calibrated CDF. The ECDF of our DKAFT models is closer to the ideal CDF than MC Dropout. Such observation is also verified by the lower values of the respective CRPS score. Within DKAFT models, the ones with an SVGP output layer perform slightly better than those with PPGP and ExactGP output layers in the PFS-PRAEGNANT prediction task. In contrast, the DKAFT models with a PPGP output layer outperform the ones with an SVGP output layer in the LoS-MIMIC prediction task.

It is worth highlighting that the inference in the MC Dropout requires multiple stochastic forward passes through the sampled network, which results in significantly slower processing than in our DKAFT models. This would further motivate the application of the analytical solutions from the proposed DKAFT models when computational efficiency plays an essential role in real-world applications, like in real-time response systems.

\section{Conclusion and Future Works}\label{sec: conclusions}

In this work, we propose the Deep Kernel Accelerated Failure Time (DKAFT) model to address the lack of uncertainty estimates in recurrent neural network (RNN) based solutions for time-to-event prediction tasks. Our DKAFT model consists of an RNN encoder and a sparse GP as the prediction model. The former serves as a trainable feature extractor to embed the patient features into a latent space of abstract covariates. The GP-based output layer consumes the abstract covariates of the patients, and outputs a predictive distribution for the time-to-event prediction.

We show that the proposed model can be trained in an end-to-end fashion, like typical neural networks, using stochastic gradient descent-based methods. In addition, a deep metric learning-based pre-training method is proposed to further improve the performance of the proposed model. Through experiments on two real-world datasets, the DKAFT models show better performance in terms of the point estimates than the RNN-based models with linear output layers. More importantly, the predictive variances from our DKAFT model reflect the confidence of the predictions. It produces better metrics evaluation in terms of RMSE and MAD with monotonically higher confidence about the predictions. Such uncertainty estimates would improve the trustworthiness of the provided model when it interacts with the physicians. Furthermore, the predictive variance also serves to improve the calibration of the model. Compared to MC Dropout, a popular method to augment the uncertainty in the neural networks, our DKAFT model shows better performance and enjoys lower computational cost, which motivates its usage in real-world applications.

As future work, we would like to further study the interpretation of the uncertainty in the proposed model. From a machine learning's perspective, it refers to checking whether a test sample lies far from the manifold constituted by the training samples.
From a decision support's perspective---since our proposed model offers predictive variances based on the ``neighboring'' training samples in the feature spaces---it would offer practical help if it also fits physicians' understanding.

\paragraph{Limitations}

As shown in Sec.~\ref{sec: svgp_tp}, in our DKAFT models, (right) censored observations will contribute to the training objective through its survival function instead of the probability density function. However, due to the nature of the time-to-event prediction tasks we focus on in this work, administrative censoring is not included in the experiments. Besides, the evaluation of censoring cases is also beyond the scope of this manuscript. We leave these perspectives as part of our future work.

\paragraph{Acknowledgement}
The authors acknowledge the support by the German Federal Ministry for Education and Research (BMBF), funding project “MLWin” (grant 01IS18050).
\begin{figure}[h]
    \includegraphics[width=0.5\linewidth]{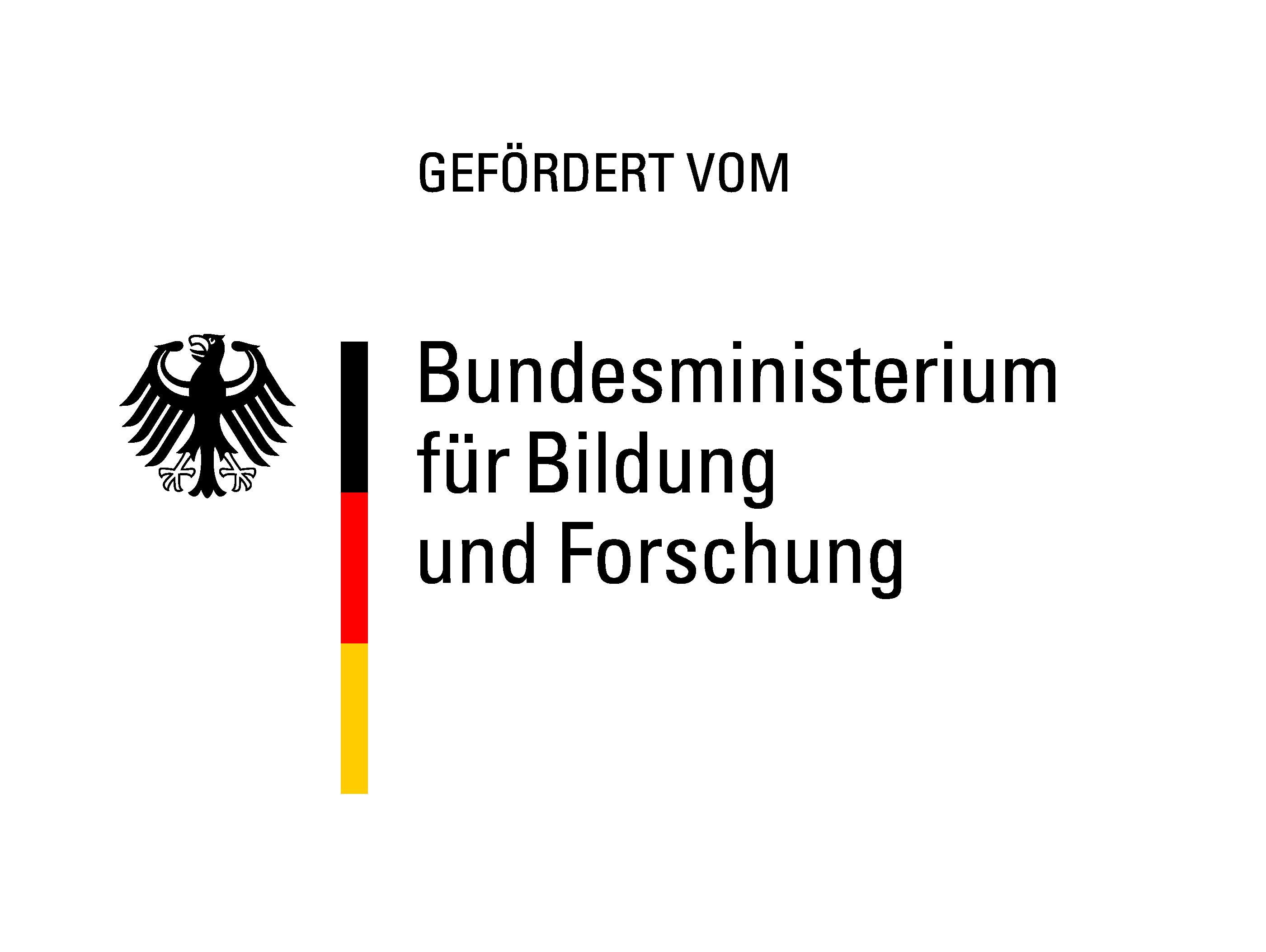}
\end{figure}

\bibliography{Ref_MLHC2021}
\newpage
\appendix

\section{Feature Set in MIMIC-III}\label{app: features}

To best represent the clinical status, we extracted both static and sequential information from the MIMIC-III. The included features are the same as the \textit{Feature Set C} defined in \citet{purushotham2018benchmarking}. In the chosen features, most have continuous values except for acquired immunodeficiency syndrome, hematologic malignancy, metastatic cancer, and admission type. The missing rates of each feature can be found in Table A.26 in \citet{purushotham2018benchmarking}.

\textbf{Static Information}: age, acquired immunodeficiency syndrome, hematologic malignancy, metastatic cancer, admission type

\textbf{Sequential Information}: Gastric Tube, Stool Out Stool, Urine Out Incontinent, Ultrafiltrate, Fecal Bag, Chest Tube 1, Chest Tube 2, Jackson Pratt 1, OR EBL, Pre-Admission, TF Residual, Albumin 5\%, Fresh Frozen Plasma, Lorazepam (Ativan), Calcium Gluconate, Midazolam (Versed), Phenylephrine, Furosemide (Lasix), Hydralazine, Norepinephrine, Magnesium Sulfate, Nitroglycerin, Insulin Regular, Morphine Sulfate, Potassium Chloride, Packed Red Blood Cells, Gastric Meds, D5 1/2NS, LR, Solution, Sterile Water, Piggyback, OR Crystalloid Intake, PO Intake, GT Flush, KCL (Bolus), Magnesium Sulfate (Bolus), Hematocrit, Platelet count, Hemoglobin, MCHC, MCH, MCV, Red blood cells, RDW, Chloride, Anion gap, Creatinine, Glucose, Magnesium, Calcium total, Phosphate, INR(PT), PT, PTT, Lymphocytes, Monocytes, Neutrophils, Basophils, Eosinophils, PH, Base excess, Calculated total CO2, PCO2, Specific gravity, Lactate, Alanine aminotransferase (ALT), Asparate aminotransferase (AST), Alkaline phosphatase, ALBUMIN, Aspirin, Bisacodyl, Docusate Sodium, Humulin-R Insulin, Metoprolol Tartrate, Pantoprazolel, Arterial Blood Pressure diastolic, Arterial Blood Pressure mean, Respiratory Rate, Alarms On, Minute Volume Alarm-Low, Peakinsp.Pressure, PEEP set, Minute Volume, Tidal Volume (observed), Minute Volume Alarm High, Mean Airway Pressure, Central Venous Pressure, Respiratory Rate (Set), Pulmonary Artery Pressure mean, O2Flow, Glucose fingerstick, Heart Rate Alarm Low, Pulmonary Artery Pressure systolic, Tidal Volume (set), Pulmonary Artery Pressure diastolic, SpO2 Desat Limit, Resp Alarm High, Skin Care, gcsverbal, gcsmotor, gcseyes, systolic blood pressure abp mean, heart rate, body temperature, pao2, fiO2, urinary output sum, serum urea nitrogen level, white blood cells count mean, serum bicarbonate level mean, sodium level mean, potassium level mean, bilirubin level, ie ratio mean, diastolic blood pressure mean, arterial pressure mean, respiratory rate, SpO2 peripheral, glucose, weight, height, hgb, platelet, chloride, creatinine, norepinephrine, epinephrine, phenylephrine, vasopressin, dopamine, midazolam, fentanyl, propofol, peep, ph.

\section{ANOVA as an ablation study}\label{app: anova}
\begin{figure}
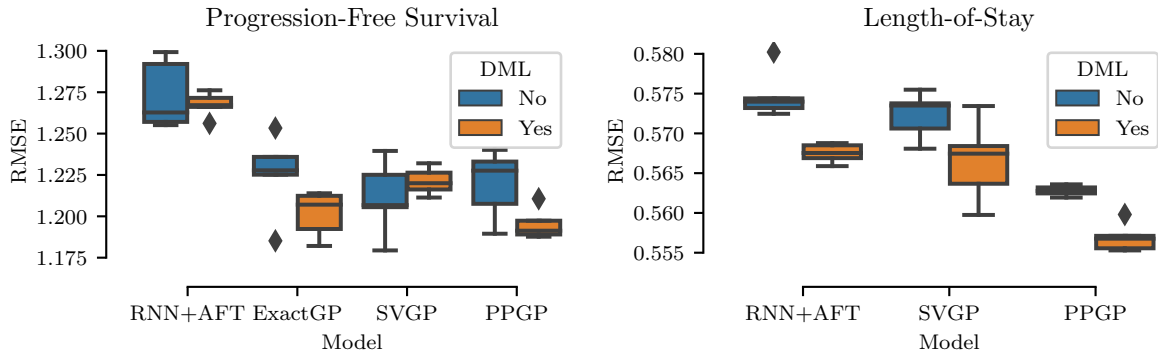

  \begin{minipage}{0.49\textwidth}
    \centering
    \input{anova_pfs.pgf}
  \end{minipage}
  \quad
  \begin{minipage}{0.49\textwidth}
    \centering
    \input{anova_los.pgf}
  \end{minipage}
    \caption{Grouped Box plots visualizing a comparison between different models, RNN+AFT and ExactGP, SVGP, PPGP of our DKAFT models. A similar conclusion to the reported $p$-values from ANOVA could be drawn from these plots. }
    \label{fig:anova}
\end{figure}

We perform ANOVA to further verify the improvements reported in Tab. \ref{tab: exp_res} in terms of RMSE. In case of the progression-free survival (PFS-PRAEGNANT) prediction task (with \citet{fasching2015biomarker} dataset), we report a $p$-value of $0.064$ w.r.t. the four choices of the prediction model, and $p$-value of $5.9\text{e-}6$ w.r.t. applying deep metric learning as pre-training or not. In case of the length-of-stay (LoS-MIMIC) prediction task (with \citet{johnson2016mimic} dataset), we report a $p$-value of $1.5\text{e-}6$ w.r.t. the three choices of the prediction model, and $p$-value of $2.0\text{e-}9$ w.r.t. applying deep metric learning as pre-training or not. It appears that only in the case of PFS-PRAEGNANT, the choice of the prediction model does not have a significant impact on the performance, presumably due to the relatively small number of patient samples.
In Fig. \ref{fig:anova} we visualize the effects of these two factors as grouped box plots.

\section{Experimental Results with More Baselines and Metrics}\label{app: baselines}
We conducted experiments for NN-based CPH models using the PyCox package\footnote{\url{https://github.com/havakv/pycox}}. Both continuous-time models (\emph{DeepSurv, CoxTime, CoxCC, and PCHazard}) and discrete-time models (\emph{LogisticHazard, PMF, DeepHit, MTLR, and BCESurv}) are included, where the latter perform much worse w.r.t. the metrics we are interested in this manuscript. Therefore, we only report the results for continuous-time models. Besides, we include two neural network architectures for these models, where a Multiple-Layer Perceptron (MLP) receives aggregated features like Cox Regression and a Recurrent Neural Network (RNN) refers to the same base network we used for our DKAFT models. In addition, we report the concordance index (C-Index) \citep{harrell1982evaluating} of all methods to show their respective discriminative performance. From Tab.~\ref{tab: exp_all} we can see, that these continuous-time models only show comparable performance to our strong baseline (RNN+AFT). Meanwhile, the RNN variants of the same model always improve the performance w.r.t. RMSE and C-Index as they take the time-dependency of the patient information into consideration.

\begin{sidewaystable}[ht]
\caption{Experimental results: MAD in the original scale (days) and RMSE in the logarithmic scale for PFS-PRAEGNANT and LoS-MIMIC prediction tasks. Our DKAFT models outperform the baselines, including Cox Regression, AFT regression, NN-based CPH models, and RNN+AFT models.}\label{tab: exp_all}
\begin{minipage}{\textwidth}
\renewcommand\footnoterule{}
\small
\begin{tabular}{cccccccc}
\toprule
                &             & \multicolumn{3}{c}{Progression-Free Survival}                                        & \multicolumn{3}{c}{Length-of-Stay}                                                 \\
Method          & Pretraining & MAD                          & RMSE                       & C-Index                  & MAD                        & RMSE                       & C-Index                  \\ \midrule
Cox Regression  & $-$\footnote{Not applicable to the method.}         & $200.800 \pm 16.984$         & $1.609 \pm 0.054$          &  $0.676\pm0.004$     & $2.727 \pm 0.007$          & $0.638 \pm 0.0002$         &     $0.683\pm0.001$    \\
AFT Regression  & $-^*$         & $206.065 \pm 8.988$          & $1.685 \pm 0.080$          &  $0.656\pm0.008$        & $2.742 \pm 0.014$          & $0.630 \pm 0.0001$         &   $0.684\pm0.001$     \\
MLP+DeepSurv    & $-^*$         & $153.900\pm7.151$            & $1.293\pm0.024$            & $0.731\pm0.006$          & $2.486\pm0.028$            & $0.593\pm0.005$            & $0.715\pm0.003$          \\
RNN+DeepSurv    & $-^*$         & $166.700 \pm 6.565$          & $1.252 \pm 0.016$          & $0.733 \pm 0.005$        & $2.456 \pm 0.039$          & $0.579 \pm 0.003$          & $0.721 \pm 0.003$        \\
MLP+CoxTime     & $-^*$         & $153.900\pm6.924$            & $1.387\pm0.018$            & $0.697\pm0.004$          & $2.599\pm0.034$            & $0.606\pm0.008$            & $0.703\pm0.005$          \\
RNN+CoxTime     & $-^*$         & $146.000 \pm 16.634$         & $1.278 \pm 0.010$          & $0.723 \pm 0.002$        & $2.515 \pm 0.051$          & $0.585 \pm 0.004$          & $0.718 \pm 0.003$        \\
MLP+CoxCC       & $-^*$         & $153.100\pm9.557$            & $1.305\pm0.010$            & $0.729\pm0.003$          & $2.463\pm0.015$            & $0.585\pm0.005$            & $0.719\pm0.003$          \\
RNN+CoxCC       & $-^*$         & $153.800 \pm 6.022$          & $1.262 \pm 0.009$          & $0.728 \pm 0.001$        & $2.479 \pm 0.018$          & $0.578 \pm 0.002$          & $0.724 \pm 0.002$        \\
MLP+PCHazard    & $-^*$         & $167.580\pm17.490$           & $1.577\pm0.189$            & $0.694\pm0.018$          & $3.036\pm0.019$            & $1.117\pm0.020$            & $0.665\pm0.006$          \\
RNN+PCHazard    & $-^*$        & $141.041 \pm 8.821$          & $1.271 \pm 0.017$          & $0.710 \pm 0.007$        & $3.055 \pm 0.021$          & $0.593 \pm 0.009$          & $0.723 \pm 0.002$        \\
RNN+AFT         & None        & $150.918 \pm 3.009$          & $1.273 \pm 0.019$          & $0.729 \pm 0.008$        & $2.476 \pm 0.040$          & $0.575 \pm 0.003$          & $0.725\pm0.002$          \\
DKAFT (ExactGP) & None        & $144.622 \pm 8.689$          & $1.225 \pm 0.022$          & $0.738\pm 0.008$         & $-$\footnote{Not possible due to the $\mathcal{O}(n^3)$ computational complexity.}                         & $-^\dagger$                       & $-^\dagger$                     \\
DKAFT (SVGP)    & None        & $154.237 \pm 13.490$         & $1.211 \pm 0.020$          & $0.747\pm0.007$          & $2.428 \pm 0.056$          & $0.572 \pm 0.003$          & $0.727\pm0.002$          \\
DKAFT (PPGP)    & None        & $147.108 \pm 6.284$          & $1.220 \pm 0.019$          & $0.745\pm0.006$          & $2.351 \pm 0.021$          & $0.563 \pm 0.001$          & $0.734\pm0.001$          \\
RNN+AFT         & DML         & $138.155 \pm 7.496$          & $1.267\pm 0.007$           & $0.732\pm0.003$          & $2.452 \pm 0.057$          & $0.568 \pm 0.001$          & $0.732\pm0.001$          \\
DKAFT (ExactGP) & DML         & $\mathbf{134.422 \pm 7.255}$ & $1.202 \pm 0.012$          & $\mathbf{0.752\pm0.003}$ & $-^\dagger$                        & $-^\dagger$                         & $-^\dagger$                      \\
DKAFT (SVGP)    & DML         & $151.852 \pm 11.305$         & $1.221 \pm 0.007$          & $0.747\pm0.002$          & $2.438 \pm 0.079$          & $0.567 \pm 0.005$          & $0.733\pm0.003$          \\
DKAFT (PPGP)    & DML         & $146.616 \pm 17.109$         & $\mathbf{1.195 \pm 0.008}$ & $\mathbf{0.752\pm0.006}$ & $\mathbf{2.346 \pm 0.042}$ & $\mathbf{0.557 \pm 0.002}$ & $\mathbf{0.738\pm0.001}$ \\ \bottomrule
\end{tabular}
\end{minipage}
\end{sidewaystable}

\end{document}